\documentclass{article} 
\usepackage{iclr2026_conference,times}

\usepackage{amsmath,amsfonts,bm}

\def\eqref#1{equation~\ref{#1}}

\def\1{\bm{1}}

\DeclareMathAlphabet{\mathsfit}{\encodingdefault}{\sfdefault}{m}{sl}
\SetMathAlphabet{\mathsfit}{bold}{\encodingdefault}{\sfdefault}{bx}{n}

\usepackage{hyperref}
\usepackage{url}
\usepackage{booktabs}
\usepackage{multirow} 
\usepackage{multicol}
\usepackage{subcaption}
\usepackage{longtable}
\usepackage{booktabs}
\usepackage{makecell}
\usepackage{graphicx}
\usepackage{subcaption}
\usepackage{wrapfig} 
\usepackage{amssymb}
\usepackage{pifont}
\usepackage{hhline}
\usepackage{svg}
\usepackage[section]{placeins}
\newcommand{\equalcontrib}{\thanks{Equal contribution as first authors.}}
\title{Exposing Hallucinations to Suppress Them: VLMs Representation Editing with Generative\\ Anchors}

\author{Youxu Shi \equalcontrib \\
University of Science and Technology of China\\
\texttt{syx123@mail.ustc.edu.cn} \\
\And
Suorong Yang \footnotemark[1] \\
Nanjing University\\
\texttt{sryang@smail.nju.edu.cn}
\AND
Dong Liu \\
University of Science and Technology of China \\
\texttt{dongeliu@ustc.edu.cn}
}

\iclrfinalcopy 
\begin{document}

\maketitle

\begin{abstract}
Multimodal large language models (MLLMs) have achieved remarkable success across diverse vision-language tasks, yet they remain highly susceptible to hallucinations, producing content that is fluent but inconsistent with visual evidence. 
Such hallucinations, spanning objects, attributes, and relations, persist even in larger models, while existing mitigation approaches often require additional fine-tuning, handcrafted priors, or trade-offs that compromise informativeness and scalability.
To address this limitation, we propose a training-free, self-supervised method for hallucination mitigation. 
Our approach introduces a novel hallucination amplification mechanism: a caption is projected into the visual space via a text-to-image model to reveal implicit hallucination signals, serving as a negative anchor, while the original image provides a positive anchor. 
Leveraging these dual anchors, we edit decoder hidden states by pulling representations toward faithful semantics and pushing them away from hallucination directions.
This correction requires no human priors or additional training costs, ensuring both effectiveness and efficiency.
Extensive experiments across multiple benchmarks show that our method significantly reduces hallucinations at the object, attribute, and relation levels while largely preserving recall and caption richness, e.g., achieving a hallucination reduction by over 5\% using LLaVA-v1.5-7B on CHAIR.
Furthermore, results on diverse architectures, including LLaVA-NEXT-7B, Cambrian-8B, and InstructBLIP-7B, validate strong cross-architecture generalization.
More importantly, when applied to hallucination-free captions, our method introduces almost no side effects, underscoring its robustness and practical plug-and-play applicability. The implementation will be publicly available.
\end{abstract}

\section{Introduction}
Multimodal Large Language Models (MLLMs)~\citep{gpt4v, zhu2025internvl3exploringadvancedtraining, geminiteam2025geminifamilyhighlycapable, bai2025qwen25vltechnicalreport, liu2023llava, vteam2025glm45vglm41vthinkingversatilemultimodal, lu2025ovis25technicalreport} 
have achieved remarkable progress on diverse vision-language tasks, including image caption~\citep{ge2024visualfactcheckerenabling, wang2023captionanythinginteractiveimage}, visual question answering~\citep{lee2024visualquestionansweringinstruction, lin2025vlmassistedcontinuallearningvisual}, and cross-modal retrieval~\citep{bai2025bridginginformationasymmetrytextvideo,yang2024clip}.
Despite the progress, MLLMs remain vulnerable to hallucinations, i.e., generating content that is fluent and plausible but inconsistent with the visual semantics.
These hallucinations can be generally categorized into three types: object-level, attribute-level, and relation-level~\citep{bai2025hallucinationmultimodallargelanguage}.
For instance, models may often mention non-existent objects, assign incorrect attributes, or describe spurious relations between objects.



Recent studies have shown that hallucinations persist even in larger and more advanced models~\citep{rohrbach2019objecthallucinationimagecaptioning,li2023evaluatingobjecthallucinationlarge,li2023blip,jiang2025interpreting}, suggesting that scaling alone is insufficient.
Existing efforts to mitigate hallucinations in MLLMs often involve external interventions (e.g., object detector)~\citep{yin2024woodpecker}, additional fine-tuning~\citep{zhang2024reflective,zhou2023analyzing, hu2024mitigatinglargelanguagemodel}, or latent editing and decoding adjustments~\citep{jiang2025interpreting, liang2024mitigatinghallucinationvisuallanguagemodels}.
However, these approaches face limitations: they rely on hand-crafted priors, fine-tuning demands additional data and computational overheads, and many latent editing-based methods often struggle with subtle hallucinations and may inadvertently suppress useful semantics.
Therefore, this raises a critical question: \textit{Can we design a training-free and self-supervised method that requires no extra supervision to mitigate hallucination without sacrificing informativeness?}

\begin{figure}[]
    \centering
    \includegraphics[width=.9\textwidth]{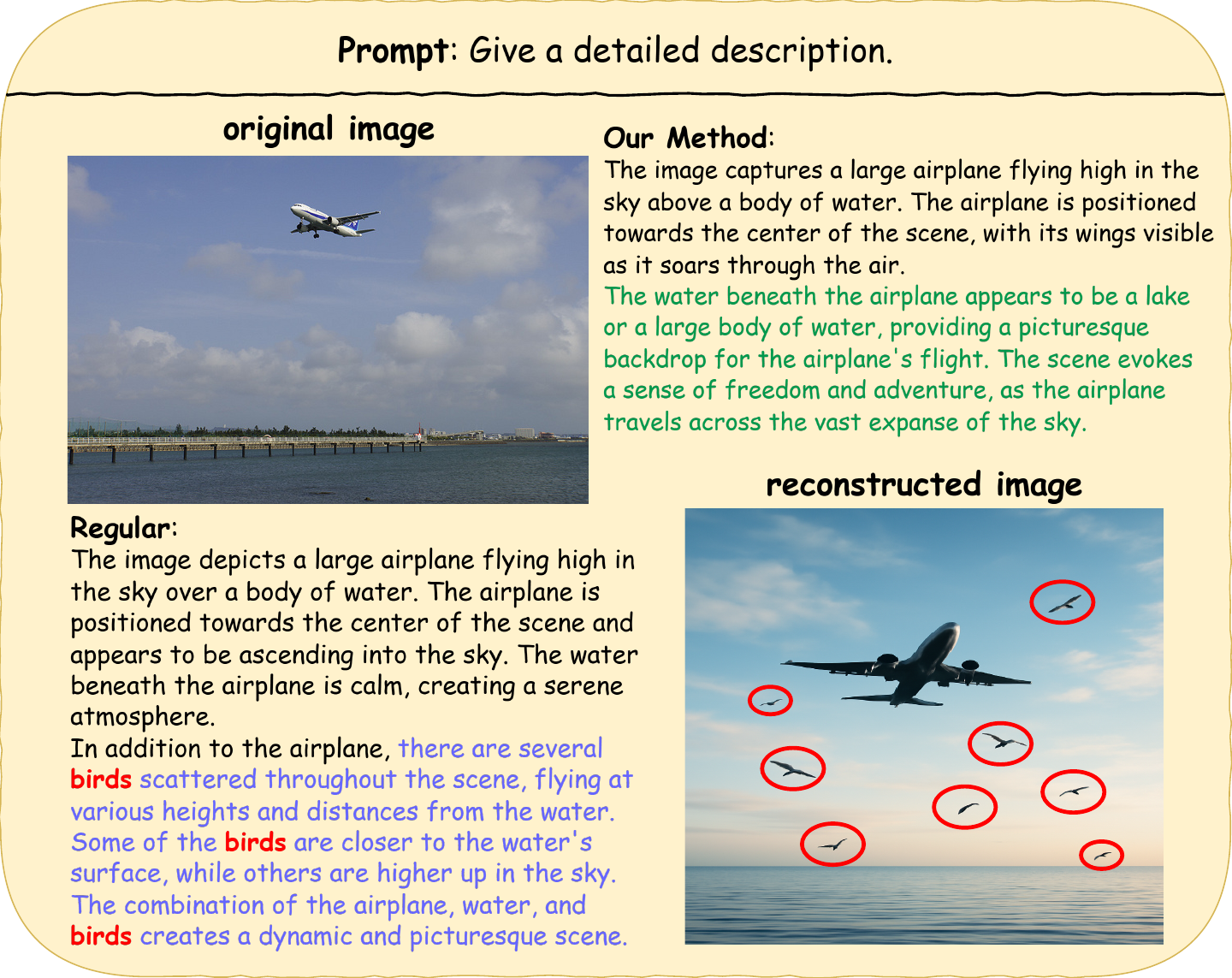}
    \caption{Illustration of our proposed method on LLaVA-v1.5-7b. The reconstructed image from the regular model's caption contains clearer hallucinated objects (circled in {\color{red}{red}}), with the corresponding hallucinated terms highlighted in {\color{red}\textbf{bold red}} with the caption. In contrast, our method not only effectively removes hallucinations but also preserves the information richness of the caption, with more accurate and specific descriptions.}
    \label{fig:case_study}
    \vspace{-5mm}
\end{figure}

To address this, we propose a training-free method that identifies correction directions to edit hidden states in the decoder layers in a self-supervised manner, after which the model produces captions that remain informative while being more faithfully grounded in visual evidence.
Specifically, since hallucinations are typically subtle and even undetectable based solely on text modality, we find that reconstructing an image from a potentially hallucinated caption via a Text-to-Image (T2I) model~\citep{ho2020denoisingdiffusionprobabilisticmodels} can expose implicit hallucination information, serving as one visual anchor during inference. 
Meanwhile, the original image provides the second visual anchor that guides the representation toward the ground truth semantics. 
Based on these dual anchors, we manipulate the latent representation by pulling image token embeddings toward the clean semantics of the original image and pushing them away from the hallucination direction derived from the reconstruction. 
In this way, our method establishes a fully self-supervised correction mechanism, requiring no human priors or additional training, while maintaining informativeness and robustness.  More importantly, when applied to hallucination-free captions, our method introduces almost no side effects, enabling seamless integration into existing MLLMs to reduce potential hallucinations without first detecting hallucinations.
Experiment results across various benchmarks and models demonstrate that our approach can effectively reduce hallucinations at the object, attribute, and relation levels while largely preserving recall and caption richness.
Moreover, results across different MLLM architectures, e.g., LLaVA-v1.5-7B~\citep{liu2023llava}, LLaVA-NEXT-7B~\citep{liu2024llavanext}, etc., demonstrate strong cross-architecture generalization and plug-and-play applicability. 

The contributions of this work can be summarized as follows: \textbf{1)} We propose a training-free, self-supervised method to mitigate hallucinations in MLLMs. By deriving supervision directly from the model's own outputs, our approach works in an entirely end-to-end and plug-and-play fashion. \textbf{2)} We introduce a novel hallucination amplification mechanism that projects caption semantics into the visual space using a T2I model. This makes otherwise implicit hallucinations perceptible and provides a lightweight way to construct reliable supervisory signals. \textbf{3)} Our method jointly anchors semantics from the original images and suppresses hallucination directions from the reconstructed image. This dual guidance removes only hallucination components while preserving genuine semantics, striking a balance between faithfulness and informativeness. \textbf{4)} Experiment results show that our method outperforms existing approaches by a large margin. Meanwhile, our approach achieves the optimal trade-off between hallucination reduction and information richness, establishing a strong baseline for future research on hallucination reduction.

\section{Related Work}
\noindent \textbf{Multi-Modal Large Language Models.}
The emergence of MLLMs~\citep{geminiteam2025geminifamilyhighlycapable, zhu2025internvl3exploringadvancedtraining} marks a significant advancement in Visual Question Answering (VQA), image captioning, and so on, extending the capabilities of traditional Large Language Models (LLMs) to process and reason across diverse modalities. MLLMs fuse visual encoders, visual projectors, and LLMs, which could leverage visual components to let MLLMs understand and reason the information mixed with image and text. For instance, GPT-4v~\citep{gpt4v} builds upon GPT-4~\citep{openai2024gpt4technicalreport}, Qwen2.5-VL~\citep{bai2025qwen25vltechnicalreport} is based on Qwen2.5-LM~\citep{qwen2025qwen25technicalreport}, and LLaVA~\citep{liu2023llava} incorporates Vicuna~\citep{vicuna2023}. Most MLLMs follow a two-stage training paradigm, consisting of pre-training and post-training. The pre-training stage exposes the model to large-scale image-text data to learn general visual knowledge. Post-training then applies refined techniques, such as Supervised Fine-Tuning (SFT)~\citep{dai2023instructblipgeneralpurposevisionlanguagemodels} and reinforcement learning (e.g., RLHF ~\citep{ouyang2022traininglanguagemodelsfollow, schulman2017proximalpolicyoptimizationalgorithms, deepseek-math}), to improve downstream performance and better align with human preferences.

\noindent \textbf{Mitigating Hallucinations in MLLMs.}
Efforts to mitigate hallucinations in MLLM can be categorized into two main directions: training-related and inference-related work~\citep{bai2025hallucinationmultimodallargelanguage}.
Training-related strategies primarily involve auxiliary supervision, which uses visual signals~\citep{chen2023mitigatinghallucinationvisuallanguage} or leveraging contrastive learning~\citep{sarkar2025mitigatingobjecthallucinationmllms}, and reinforcement learning from human feedback~\citep{benkish2024mitigatingopenvocabularycaptionhallucinations, sun2023aligninglargemultimodalmodels, yu2024rlhfvtrustworthymllmsbehavior} to obtain more reliable and trustworthy model outputs. Inference-related approaches offer lightweight and efficient alternatives that do not require retraining the model. Many of them put efforts into the decoding strategy, such as Contrastive Decoding VCD~\citep{leng2023mitigatingobjecthallucinationslarge},  ICD~\citep{wang2024mitigatinghallucinationslargevisionlanguage}, and Guided Decoding DeCo~\citep{wang2025mllmseedynamiccorrection}. Some studies also explore modifying intermediate representations within the language model component of VLMs, which not only helps mitigate hallucinations but also offers a good interpretable insight~\citep{jiang2025interpreting, liu2024reducinghallucinationsvisionlanguagemodels}. A great body of work has shown that these training-free or post-hoc revision methods can achieve or even outperform training-related methods~\citep{ge2024visualfactcheckerenabling}, making them appealing for practical applications.

\noindent \textbf{Text-Image Generative Model Efforts in VLMs.}
Text-to-Image (T2I) generative models have been increasingly integrated into the development and evaluation of MLLMs, typically for two major purposes. First, T2I models are used for benchmark construction, where generated images are based on corresponding captions that include diverse open-vocabulary objects using a T2I diffusion model~\citep{benkish2024mitigatingopenvocabularycaptionhallucinations}.
Second, T2I models play a crucial role in the construction of training datasets. V-DPO~\citep{xie2024vdpomitigatinghallucinationlarge} devises a vision-guided direct preference optimization with a synthetic dataset containing both response-contrast and image-contrast preference pairs. ESREAL~\citep{kim2024esrealexploitingsemanticreconstruction} proposes to utilize a T2I model to semantically reconstruct an image from the generated caption. 
After that, the semantic misalignment between the two images can serve as feedback to optimize the model.

\section{Proposed Method}
\subsection{Overview}
\begin{figure}[]
    \centering
    \includegraphics[width=1.0\textwidth]{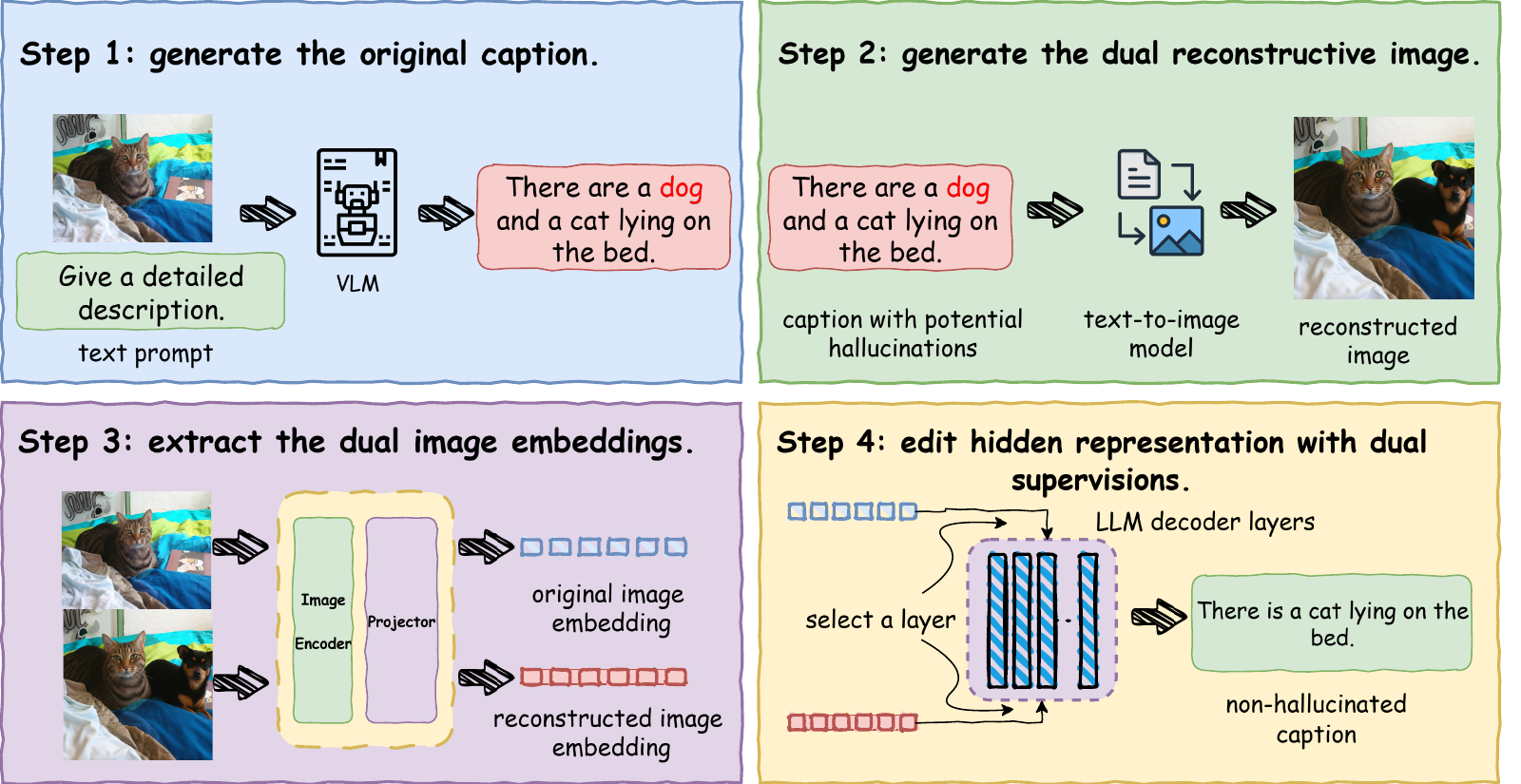}
    \caption{Overview of the proposed method. Given an image and a query, the MLLM generates a caption that may contain hallucinations. The caption is then fed into a T2I model to reconstruct an auxiliary image, which amplifies potential hallucinations.
    Next, both the original and reconstructed images are encoded into embeddings that serve as dual anchors. By injecting these embeddings into the decoder layers to edit hidden representations during inference, the model produces captions that are faithful to visual content without sacrificing informativeness.}
    \label{fig:pipeline} 
    \vspace{-5mm}
\end{figure}

As illustrated in Fig.\ref{fig:pipeline}, our method proposes an end-to-end pipeline to mitigate hallucination.
Given an input image, we use a VLM to produce an initial caption, which may include hallucinated objects or relations. 
To expose the potential hallucinations in captions, we synthesize a reconstructed image based on the caption using a text-to-image (T2I) model.
This reconstruction can naturally amplify and externalize hallucinated content into the visual space.
As a result, hallucinations that were originally implicit and difficult to detect in textual semantic space become perceptible once projected into images.
Both the original image and the reconstructed image are fed through the image encoder and projection head to obtain the embeddings, denoted as $f(\boldsymbol{I})$ and $f(\boldsymbol{I}')$, respectively.
Here, $f(\boldsymbol{I})$ acts as a clean semantic anchor, guiding the representation toward faithful visual semantics, while $f(\boldsymbol{I}')$ explicitly captures the hallucination direction amplified through reconstruction.
By simultaneously pulling the image token embeddings toward $f(\boldsymbol{I})$ and pushing them away from $f(\boldsymbol{I}')$, our method establishes an adversarial correction mechanism that requires no hand-crafted metrics or external supervision.
Therefore, this design transforms hallucination suppression into a fully self-supervised process, enabling end-to-end correction without human intervention.

\begin{wrapfigure}[12]{l}{0.45\linewidth}
\centering
\vspace{-13pt}
    \includegraphics[width=0.35\textwidth]{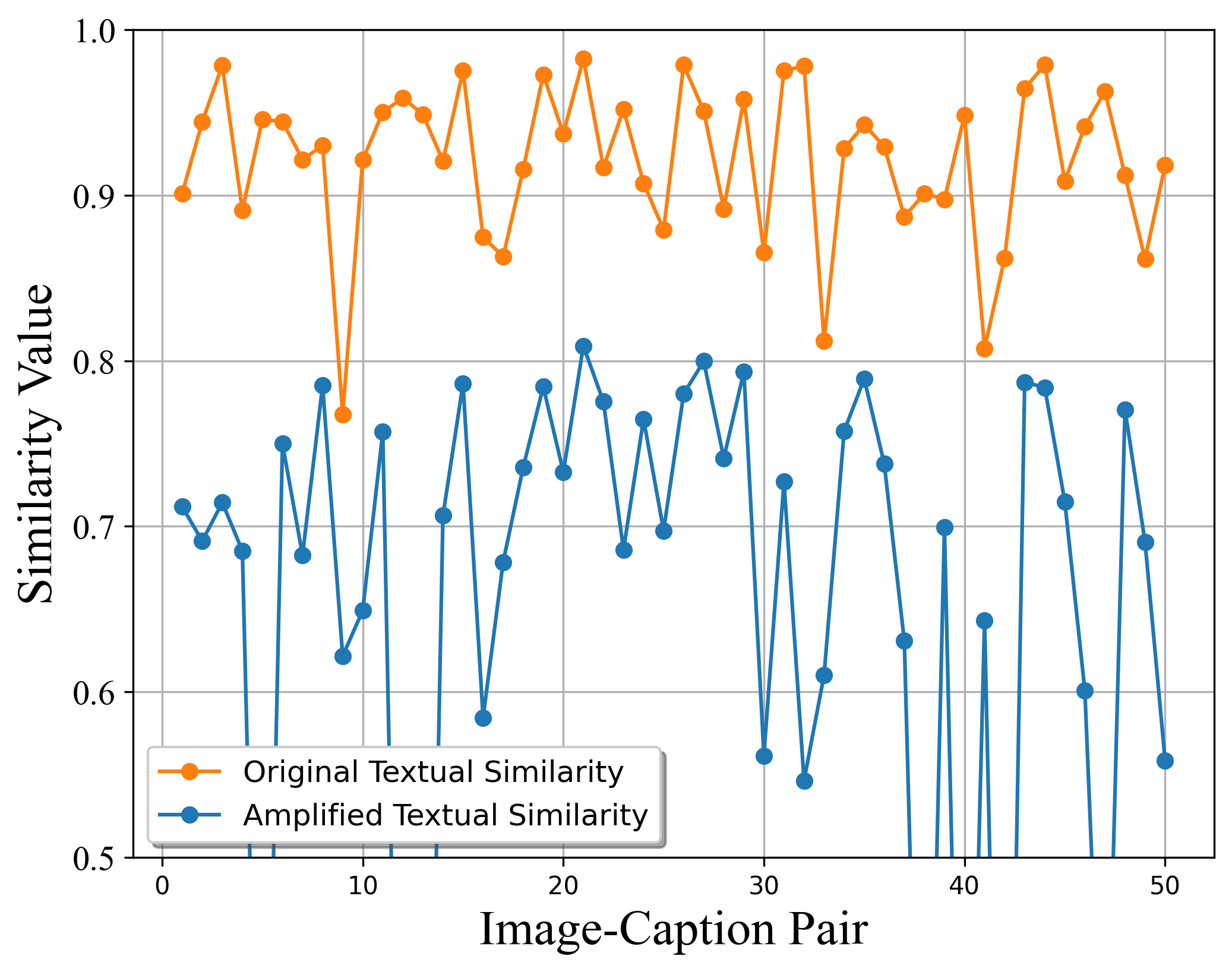}
    \caption{Illustration of the effectiveness of the hallucination amplification mechanism.}
    \label{fig:simi} 
\end{wrapfigure}
\subsection{How to Expose Hallucinations}
Hallucinations in MLLMs are intrinsically difficult to detect because they are linguistically well-formed and often indistinguishable from faithful descriptions at the text level.
The discrepancy lies not in language plausibility but in the misalignment with visual evidence, which the model itself is typically insensitive to.
To address this, we introduce a hallucination exposure mechanism that leverages generative reconstruction to convert implicit inconsistencies into explicit and observable signals.
Specifically, given an input image and its generated caption, we use a text-to-image (T2I) model, e.g., FLUX.1-dev~\citep{labs2025flux1kontextflowmatching, flux2024}, to reconstruct an image from the caption.
This reconstructed image exaggerates and materializes the semantics, including hallucinated objects or relations, thereby amplifying otherwise hidden inconsistencies and providing usable supervisory signals for subsequent correction.


To further validate this intuition, Fig.\ref{fig:simi} illustrates the effectiveness of our hallucination amplification mechanism.
For consistency in semantic representation, we compute similarity purely in the textual space. Concretely, let the original image and its caption be denoted as $\boldsymbol{I}$ and $\boldsymbol{\tau}$. By injecting hallucinated information into $\boldsymbol{\tau}$, we obtain a hallucinated caption $\boldsymbol{\tau}'$, which is then fed into a T2I model to reconstruct an image $\boldsymbol{I}'$.
Instead of directly comparing the embeddings of $\boldsymbol{I}$ and $\boldsymbol{I}'$, we leverage LLaVA to caption both images, producing $\boldsymbol{t}$ and $\boldsymbol{t}'$, and compute the similarity between them.
Thus, we compare the differences between $sim(\boldsymbol{\tau},\boldsymbol{\tau}')$ and $sim(\boldsymbol{t},\boldsymbol{t}')$.
Notably, even though both the similarity is measured between captions in the textual space, we observe a significant drop once hallucinations are introduced and amplified via reconstruction.
This demonstrates that our amplification mechanism transforms subtle, implicit hallucinations into detectable semantic deviations.
More importantly, the proposed cross-modal amplification mechanism is entirely training-free and can be applied in a post-hoc manner. 
It reveals a broader spectrum of hallucination types, including relational and logical inconsistencies, that remain subtle in pure text space, providing an efficient and generalizable pathway.

\subsection{Hallucination Mitigation via Latent Editing}
\noindent \textbf{Dual Supervision Construction.}
We leverage cross-modality reconstruction to amplify potential hallucinations, turning subtle semantic noise in captions into perceptible signals in the visual space.
Since our ultimate goal is to mitigate hallucinations in caption generation, it is more effective to operate on the image-space latent representations. 
We use the original input image $\boldsymbol{I}$ as a semantic anchor, which encodes the clean and faithful semantics.
Its embedding, denoted as $f(\boldsymbol{I})$, obtained through the image encoder and projection head, naturally serves as a ground-truth supervision signal. By pulling the latent representation of the caption closer to this anchor, we ensure that editing does not distort or erase genuine visual information.
Meanwhile, the reconstructed image $\boldsymbol{I}'$ is generated based on the obtained caption, which may originally contain hallucinated information. 
Its embedding $f(\boldsymbol{I}')$ thus provides potentially negative supervision signals, pointing to the direction in the latent space that corresponds to potential hallucinations.
By pushing the caption representation away from this embedding, we discourage it from retaining hallucination-related features.

\noindent \textbf{Latent Representation Editing.}
Previous studies~\citep{huh2024platonicrepresentationhypothesis,jiang2025interpreting,liu2025reducing,stolfo2025improvinginstructionfollowinglanguagemodels} have shown that steering and editing latent representation can guide the generation of LLMs, and have been applied to various tasks such as instruction following and hallucination mitigation, with correction directions typically from handcrafted priors or training objectives.
In contrast, our proposed method introduces a training-free and dual-supervision scheme.
We combine these two directions and edit the image tokens in the embedding extracted from a selected decoder layer as follows:
\begin{equation}\label{eq:dual-supervision}
K_{h, l}^\prime = K_{h, l} + \alpha f(I) - \beta f(I^\prime),\; h \in \mathcal{H}_{img},\; l \in [1, L],
\end{equation}
where $K_{h, l}$ stands for the embedding of the $h$-th token at the $l$-th decoder layer, $\mathcal{H}_{img}=\{i_1, i_1+1, \dots i_n\}$ represents the set of positions corresponding to the image-related tokens, $\alpha$ and $\beta$ are scalar coefficients controlling the contributions from the original image $I$ and the dual image $I^\prime$, respectively, and $f(\cdot)$ denotes the combined transformation of image encoder and projector.

Since the potential hallucination signals are derived directly from the model’s outputs without requiring external annotation, and the supervision employed for editing is also automatically constructed within the same pipeline, the entire approach constitutes a fully self-supervised, end-to-end knowledge editing strategy that requires no human intervention.
Thus, our method can be used as a plug-and-play module in the VLM inference process.


\section{Experiments}
\subsection{Experimental Setup}
\noindent \textbf{Datasets and Evaluation Metrics.}
To comprehensively validate the effectiveness of our method on mitigating hallucination issues on large vision-language models, we utilize the Caption Hallucination Assessment with Image Relevance (CHAIR)~\citep{rohrbach2019objecthallucinationimagecaptioning}, MLLM Evaluation benchmark (MME)~\citep{fu2024mmecomprehensiveevaluationbenchmark}, and Pooling-based Object Probing Evaluation (POPE)~\citep{li2023evaluatingobjecthallucinationlarge} benchmarks. 

However, directly using the hallucination rate or recall may not fully assess hallucination mitigation performance.
For example, a model can trivially achieve a very low hallucination rate by producing no or overly short captions, but such outputs fail to capture the visual content.
To address this, we combine the hallucination rate and the recall into a unified metric and therefore propose the metric $HAR_{\mathtt{@}\beta}$ (Hallucination and Recall), which is defined as:
\begin{equation}
\begin{aligned}
        HAR_{\mathtt{@}\beta} = \frac{(1+\beta^2rq)}{\beta^2q+r}, \; q = 1-h
\end{aligned}
\end{equation}
where $h$ denotes the hallucination rate CHAIR and $r$ denotes the recall ($HAR \in [0,1]$). The metric is monotonic in both $q$ and $r$, and the parameter $\beta$ controls the trade-off: $\beta > 1$ emphasizes recall, whereas $\beta < 1$ emphasizes reducing hallucinations. 
In this way, a high $HAR_{\beta}$ can only be achieved when both the recall and the non-hallucinated rates are high.
POPE serves as a complement to CHAIR in object hallucinations on SEEM-annotated datasets, including MSCOCO~\citep{lin2015microsoftcococommonobjects}, A-OKVQA~\citep{schwenk2022aokvqabenchmarkvisualquestion}, and GQA~\citep{hudson2019gqanewdatasetrealworld}, thereby broadening the evaluation scope by focusing on the model's performance with context-dependent prompts. 
Moreover, for hallucination at the attribute level, we adopt the MME benchmark. 
Similar to POPE, MME tasks are framed as binary Yes-or-No questions, facilitating consistent evaluation.
Some results for baseline methods are taken from~\citep{zou2025looktwiceanswermemoryspace}.

\noindent \textbf{Experiment Settings.}
We conduct experiments using several of the most representative model architectures.
Unless specified, we adopt LLaVA-v1.5-7B \citep{liu2023llava}, a widely adopted large vision-language model for image–text understanding and captioning, and FLUX.1-dev, a popular open-source generative model that exemplifies recent advances in text-to-image generation.
For fairness, in our comparison, we set all the decoding hyperparameters and temperature the same across different decoding methods.
Closely following~\citep{jiang2025interpreting}, in our method, we also conduct latent editing in the second layer and uniformly set both $\alpha$ and $\beta$ to 0.1.

\subsection{Hallucination Mitigation Results}
In this section, we evaluate the model on three benchmarks, i.e., CHAIR, MME, and POPE, to assess its performance in image captioning, object hallucination, and attribute hallucination. 

\noindent \textbf{Results on CHAIR.} 
As shown in Table~\ref{tab:chair}, our method consistently outperforms other baselines on both $\text{CHAIR}_S$ and $\text{CHAIR}_I$, demonstrating its superior effectiveness in suppressing hallucinations.
Meanwhile, although nearly all methods inevitably reduce recall while suppressing hallucinations, reflecting a trade-off between faithfulness and informativeness, our approach achieves the smallest drop. 
This demonstrates that our method captures a broad range of ground-truth objects.
With the $HAR_{\mathtt{@}\beta}$ metric, our method achieves the highest score, highlighting its ability to reduce hallucinations while maintaining coverage.
This superior performance originates from our dual-supervision construction, which simultaneously anchors clean semantics from the original image and suppresses hallucination directions from the reconstructed image.
In essence, this editing removes only the component along this direction rather than globally suppressing the representation.
Thus, our method mitigates hallucinations with little side effect, preserving both informativeness and semantic richness.

\begin{table}[]
\centering
\caption{Performance of hallucination mitigation on CHAIR across various metrics.}
\resizebox{.75\textwidth}{!}{
\begin{tabular}{l|cccccc}
\toprule
Methods & CHAIR$_S$ $\downarrow$ & CHAIR$_I$ $\downarrow$ & Average $\downarrow$ & Length$\uparrow$  & Recall $\uparrow$ & $HAR_{\mathtt{@} 1}$ $\uparrow$ \\
\midrule
Baseline & 53.0  & 14.0  & 33.50 & 92.20 &  81.00 & 0.7304\\
OPERA & \textbf{47.8} & 14.6 & 31.20 & 98.65 & 76.80 & 0.7258 \\
ICD & 56.2 & 16.3 & 36.25 & 103.40 & 16.31 & 0.2596\\
VCD & 48.7 & 14.9 & 31.80 & 100.40 & 77.32 & 0.7247\\
Ours & \textbf{47.8} & \textbf{12.7} & \textbf{30.25} & 92.44 & 80.10 & \textbf{0.7460}\\
\bottomrule
\end{tabular}
}
\label{tab:chair}
\vspace{-5pt}
\end{table}

\begin{table}[]
\centering
\caption{Performance comparisons on POPE across different settings and datasets.}
\label{tab:pope}
\resizebox{.96\columnwidth}{!}{
\begin{tabular}{@{}clcccccccccc@{}}
\toprule
\multirow{2}{*}{Dataset} & \multirow{2}{*}{Methods} & \multicolumn{2}{c}{Random} & \multicolumn{2}{c}{Popular} & \multicolumn{2}{c}{Adversarial} & \multicolumn{2}{c}{Average} \\
\cmidrule(lr){3-4} \cmidrule(lr){5-6} \cmidrule(lr){7-8} \cmidrule(lr){9-10}
& & Accuracy $\uparrow$ & F1-score $\uparrow$ & Accuracy $\uparrow$ & F1-score $\uparrow$ & Accuracy $\uparrow$ & F1-score $\uparrow$ & Accuracy $\uparrow$ & F1-score $\uparrow$ \\
\midrule
\multirow{5}{*}{MSCOCO}
& Baseline  & 83.49 & 82.28 & 79.98 & 79.34 & 76.03 & 76.26 & 79.83 & 79.29 \\
& ICD   & 84.87 & 83.27 & 82.93 & 81.45 & 81.07 & 79.96 & 82.96 & 81.56 \\
& VCD   & 86.84 & 86.83 & 82.65 & 83.37 & 77.31 & 79.28 & 82.27 & 83.16 \\
& OPERA   & 87.53  & 86.45 & 84.21  & 83.50 & 80.88 & 80.69 & 84.21 & 83.55 \\
& Ours & \textbf{89.35} & \textbf{89.43} & \textbf{86.00} & \textbf{86.14} & \textbf{81.76} & \textbf{82.84} & \textbf{85.70} & \textbf{86.14} \\
\cmidrule{1-10}
\multirow{5}{*}{A-OKVQA}
& Baseline  & 83.45 & 82.56 & 79.90 & 79.59 & 74.04 & 75.15 & 79.13 & 79.10 \\
& ICD   & 85.57 & 85.06 & 81.93 & 81.95 & 77.43 & 78.99 & 81.64 & 82.00 \\
& VCD   & 86.15 & 86.34 & 81.85 & 82.82 & 74.97 & 77.73 & 80.99 & 82.30 \\
& OPERA   & 88.27 & 87.54 & 85.17 & 84.74 & \textbf{79.37} & 79.97 & 84.27 & 84.08 \\
& Ours & \textbf{89.53} &  \textbf{89.31}  &  \textbf{86.50}  &  \textbf{86.63}  &   79.20  & \textbf{80.79}   &  \textbf{85.08}  &  \textbf{85.58}   \\
\cmidrule{1-10}
\multirow{5}{*}{GQA}
& Baseline  & 83.73 & 82.95 & 78.17 & 78.37 &75.08  & 76.06 & 78.99 & 79.13 \\
& ICD   & 84.90 & 84.22 & 78.37 & 78.81 & 75.97 & 76.93 & 79.75 & 79.99 \\
& VCD   & 86.65 & 86.99 & 80.73 & 82.24 & 76.09 & 78.78 & 81.16 & 82.67 \\
& OPERA   & 83.73 & 82.95 & 78.17 & 78.37 & 75.08 & 76.06 & 78.99 & 83.83 \\
& Ours &  \textbf{87.80}  & \textbf{87.43}   &  \textbf{82.73}  & \textbf{83.01}   &  \textbf{79.73}  &  \textbf{80.66}  &  \textbf{83.42}  &  \textbf{83.70}  \\
\bottomrule
\end{tabular}}
\vspace{-1mm}
\end{table}
\noindent \textbf{Results on POPE.} 
While CHAIR primarily evaluates caption hallucinations at the object level, POPE complements it with a polling-based framework.
The performance on the POPE benchmark under random, popular, and adversarial settings is presented in Table~\ref{tab:pope}. It can be observed that our method consistently achieves the best performance across all settings.
Notably, our method can achieve up to +5.95\% accuracy and +6.85\% F1 score on average, outperforming other training-free approaches by a large margin.
Therefore, these results demonstrate that our method provides a reliable and generalizable solution across different levels of difficulty. 

\noindent \textbf{Results on MME.}
In addition to object-level hallucinations, we further evaluate our method on the MME benchmark, which targets attribute-level hallucinations, which is a finer-grained and considerably more challenging setting.
It can be observed from Table~\ref{tab:mme} that our method achieves strong performance on the MME benchmark.
In particular, it achieves notable gains on position- and count-related tasks, while maintaining competitive results on object-level existence and color recognition.
Moreover, we observe a slight drop in color-related tasks, which can be attributed to the inherent limitations of T2I models in handling fine-grained color information.
This phenomenon suggests that the reconstructed images, while effective in amplifying and exposing hallucination directions, may introduce uncertainty in some more fine-grained perception, thereby influencing the internal knowledge of the MLLMs. We further discuss this point in Appendix~\ref{Discussion}.
\begin{figure}
\vspace{-2mm}
    \begin{minipage}{0.6\textwidth}
    \centering
\captionof{table}{Performance comparisons on MME.}
\label{tab:mme}
\resizebox{.99\linewidth}{!}{
\begin{tabular}{@{}lc|cc|cc@{}}
\toprule
\multirow{2}{*}{Methods} & 
\multirow{2}{*}{MME-Hall Total $\uparrow$} & 
\multicolumn{2}{c|}{Object-Level} & 
\multicolumn{2}{c}{Attribute-Level} \\
\cmidrule(lr){3-4} \cmidrule(lr){5-6}
& & Existence $\uparrow$ & Count $\uparrow$ & Position $\uparrow$ & Color $\uparrow$ \\ 
\midrule
Baseline & 643.3 & 190.0 & 155.0 & 128.3 & 170.0 \\
 ICD & 583.3 & 185.0 & 130.0 & 121.7 & 146.7 \\
 VCD & 648.3 & 190.0 & 155.0 & 133.3 & \textbf{170.0} \\
 OPERA & 610.0 & \textbf{195.0} & 128.3 & 121.7 & 165.0 \\
 Ours & \textbf{667.3} & 193.0 & \textbf{156.7} & \textbf{150.0} & 167.7 \\ 
\bottomrule
\end{tabular}
}
    \end{minipage}
    \hspace{5pt}
    \begin{minipage}{0.35\textwidth}
        \centering
\captionof{table}{Robustness to hallucination-free captions across different VLMs. }
\vspace{-4pt}
\renewcommand{\arraystretch}{1.2}
\resizebox{.99\textwidth}{!}{
\begin{tabular}{c | cc }
\toprule
& $\Delta$ of CHAIR &$\Delta$ of Recall \\
\midrule
{LLaVA-v1.5-7B} & 0 & -2 \\  
{LLaVA-NEXT-7B} & 0 & 0 \\
{Cambrian-8B} & 0 & -2 \\
{InstructBLIP-7B} & 0 & +1 \\
\bottomrule
\end{tabular}}
\label{tab:Robustness}
    \end{minipage}
    \vspace{-5mm}
\end{figure}


\noindent \textbf{Cross-Architecture Generalization.}
We further employ our method to other widely-used VLM, including LLaVA-NEXT-7B \citep{liu2024llavanext}, Cambrian-8B \citep{tong2024cambrian}, and InstructBLIP-7B \citep{dai2023instructblipgeneralpurposevisionlanguagemodels}.
As shown in Table~\ref{tab:cross-arch}, our approach demonstrates consistent improvements in hallucination mitigation across different architectures.
Notably, on LLaVA-NEXT-7B, our method substantially reduces hallucinations, with CHAIR$_S$ decreasing from 22.4 to 6.6 and CHAIR$_I$ from 6.0 to 2.9. 
On InstructBLIP-7B, our method not only reduces hallucinations but also improves recall compared to the baseline.
Importantly, these gains are achieved without sacrificing caption length or informativeness, striking a favorable balance between faithfulness and coverage.
Overall, the results highlight that our proposed dual-supervision editing mechanism can serve as a plug-and-play module across diverse VLMs.

\begin{table}[t]
\centering
\caption{Cross-architecture generalization of our method on LLaVA-NEXT-7B, Cambrian-8B, and InstructBLIP-7B.}
\renewcommand{\arraystretch}{1.2}
\resizebox{.8\textwidth}{!}{
\begin{tabular}{c| c cccccc}
\toprule
& & \multicolumn{5}{c}{\textbf{CHAIR}} \\
&  Method  & CHAIR$_S$ & CHAIR$_I$ & Average & Length & Recall & $HAR_{\mathtt{@} 1}$ $\uparrow$\\
\midrule
\multirow{2}{*}{LLaVA-NEXT-7B}
& Baseline & 22.4 & 6.0 & 14.20 & 167.78 & 64.40 & 0.7358 \\
& Ours & \textbf{6.6} & \textbf{2.9} & \textbf{4.75} & 119.51 & 60.21 & \textbf{0.7377} \\
\midrule
\multirow{2}{*}{Cambrian-8B}
& Baseline & 9.6 & 3.8 & 6.70 & 65.45 & 53.42 & 0.6792 \\
& Ours & \textbf{8.0} & \textbf{2.9} & \textbf{5.45} & 67.43 & 53.29 & \textbf{0.6809} \\
\midrule
\multirow{2}{*}{InstructBLIP-7B}
& Baseline & 57.4 & \textbf{15.5} & 36.45 & 98.08 & 74.70 & 0.6868\\
& Ours  & \textbf{56.4} & 15.8 & \textbf{36.10} & 98.61 & 75.17 & \textbf{0.6905}\\
\bottomrule
\end{tabular}}
\label{tab:cross-arch}
\end{table}


\noindent \textbf{Robustness Under Non-Hallucinated Conditions}
An important property of hallucination mitigation is its robustness: while reducing hallucinations is desirable, the method should not degrade captions that contain no hallucinations.
While most existing methods do not evaluate robustness on hallucination-free cases, we explicitly test this by applying our editing mechanism to captions without hallucinations and comparing the outputs before and after latent editing.
As shown in Table~\ref{tab:Robustness}, we report the average change of CHAIR and recall per caption across different models. 
The results show that the CHAIR metric remains unchanged while recall presents a slight fluctuation ($\pm2$ on average).
The results demonstrate that our editing is robust: it suppresses hallucinations when they exist, yet does not introduce side effects when hallucinations are absent.
As a result, our method can be seamlessly plugged into different VLMs without the need to first detect hallucinations, highlighting the practical significance.

\subsection{Analytical Results}
\noindent \textbf{Effect of Weight Factor $\alpha$ and $\beta$}. 
We directly set the weight factors in Eq.\ref{eq:dual-supervision} following~\citep{jiang2025interpreting}.
In this section, to investigate the influence of weight factors $\alpha$ and $\beta$ in our dual-supervision editing, we conduct experiments under both symmetric ($\alpha = \beta$) and asymmetric ($\alpha \neq \beta$) settings.
As shown in Table \ref{tab:alpga_beta}, under the symmetric setting, increasing $\alpha$ and $\beta$ progressively suppresses hallucinations. 
However, recall simultaneously drops from 80.18 to 75.98 while the response length increases.
This shows that larger weights enforce stronger hallucination suppression, but at the cost of reduced content coverage and more verbose outputs.
Moreover, the asymmetric setting reveals that encouraging negative supervision, i.e., $\beta > \alpha$, achieves a greater hallucination mitigation while the model is also overly conservative and tends to produce repetitive responses.
In contrast, when $\beta$ is smaller than $\alpha$, recall increases, but this comes at the expense of higher hallucination rates.
Therefore, the symmetric setting strikes a favorable trade-off, delivering effective hallucination suppression while preserving recall and avoiding overly conservative behavior.

\begin{table}[]
\centering
\renewcommand{\arraystretch}{1.2}
\caption{Analytical results on weight factor $\alpha$ and $\beta$ across different settings.}
\resizebox{.8\textwidth}{!}{
\begin{tabular}{c cc cccccc}
\toprule
& $\alpha$ & $\beta$ & CHAIR$_S$ & CHAIR$_I$ & Average & Length & Recall & $HAR_{\mathtt{@} 1}$ $\uparrow$\\
\midrule
\multirow{4}{*}{$\alpha = \beta$}
& 0.09 & 0.09 & 51.8 & 13.9 & 32.85 & 92.33 & 80.18 & 0.7310\\
& 0.10 & 0.10 & 47.8 & 12.7 & 30.25 & 92.44 & 80.05 & 0.7456\\
& 0.11 & 0.11 & 42.4 & 11.8 & 27.10 & 97.66 & 77.49 & 0.7511\\
& 0.12 & 0.12 & 40.2 & 12.7 & 26.45 & 97.38 & 75.98 & 0.7460\\
\midrule
\multirow{4}{*}{$\alpha \ne \beta$}
& 0.08 & 0.12 & 38.6 & 12.4 & 25.50 & 101.80 & 74.48 & 0.7448\\
& 0.09 & 0.11 & 41.0 & 11.9 & 26.45 & 100.02 & 76.90 & 0.7519\\
& 0.11 & 0.09 & 51.2 & 13.5 & 32.35 & 92.32 & 80.91 & 0.7368\\
& 0.12 & 0.08 & 52.0 & 14.0 & 33.00 & 92.33 & 81.69 & 0.7362\\
\bottomrule
\end{tabular}}
\label{tab:alpga_beta}
\end{table}


\begin{table}[t]
\centering
\renewcommand{\arraystretch}{1.2}
\caption{Ablation study of the two supervision components in our algorithm using LLaVA-1.5. }
\resizebox{.85\textwidth}{!}{
\begin{tabular}{cc cccccc}
\toprule
$+\alpha f(I)$ & $-\beta f(I^\prime)$ & CHAIR$_S$ & CHAIR$_I$ & Average & Length & Recall & $HAR_{\mathtt{@} 1}$ $\uparrow$\\
\midrule
\ding{55} & \ding{55} & 53.0 & 14.0 & 33.50 & 92.20 & 81.00 & 0.7300 \\
\checkmark & \ding{55} & 51.4 & 14.2 & 32.80 & 91.46 & 80.61 & 0.7330\\
\ding{55} & \checkmark & 46.4 & 12.0 & 29.20 & 101.42 & 76.62 & 0.7360\\
\checkmark & \checkmark & 47.8 & 12.7 & 30.25 & 92.41 & 80.05 & 0.7456\\
\bottomrule
\end{tabular}}
\label{tab:two_key}
\end{table}

\noindent \textbf{Effect of Dual Supervision Signals.}
To better understand the effect of each supervision signal in Eq.\ref{eq:dual-supervision}, we conduct ablation studies.
As shown in Table~\ref{tab:two_key}, removing both signals represents the baseline performance, which results in the highest hallucination rates.
Introducing only the negative supervision $-\beta f(I^\prime)$ presents very strong effectiveness in hallucination mitigation and increases the response length.
However, we find that in this case, the model behaves overly conservatively: a substantial portion of the responses are repetitive or focus solely on captioning a single object.
Thus, the recall is the lowest among variants.
While the response length increases, the information richness does not increase, indicating degraded captioning capability.
In contrast, introducing only the positive supervision $+\alpha f(I)$  emphasizes the visual tokens of the original image.
Since this variant does not introduce any component for hallucination mitigation, it brings limited influence on hallucination performance.
When combining both components, our method achieves the best trade-off overall, e.g., achieving the highest $HAR_{\mathtt{@}1}$ score.
Specifically, since the dual supervision signals influence model performance in different directions, the results deteriorate when either one is absent. 
For instance, the presence of $f(I')$ tends to make the model overly conservative: while it suppresses fabricated content, it also prevents the model from recognizing objects that actually exist. In contrast, the presence of $f(I)$ serves as a compensation for this drawback.
Therefore, integrating $+f(I)$ with $-f(I')$ allows the model to preserve information richness and coverage, while still ensuring effective suppression of hallucinations.
As a result, removing any component from our method inevitably leads to suboptimal performance.

\noindent{\textbf{Effect of Layer $l$}. 
Table~\ref{tab:effect-layer} shows the results of applying our method at different decoder layers.
It can be seen that our method consistently proves effective across most layers.
Consistent with the results in \citep{jiang2025interpreting}, we observe that intervening at relatively shallow layers yields stronger hallucination suppression, as the injected supervision can influence downstream representations more directly.
However, this often comes with a slight drop in recall, suggesting that shallow-layer edits may over-regularize the model’s perception.
In contrast, applying edits at middle or deeper layers better preserves recall and caption richness while still reducing hallucinations to a certain extent.
We will further discuss the layer selection in Appendix~\ref{Discussion}.

\noindent \textbf{Effect of the Generative Model.} 
To assess the effect of the hallucination amplification mechanism and the generative models, we evaluate the effectiveness of our method using a different generative model, GPT-image~\citep{gptimage}.
As shown in Table \ref{tab:effect-generative-model}, both variants substantially reduce hallucinations compared to the baseline, achieving lower CHAIR$_S$ and CHAIR$_I$ scores.
Importantly, the overall improvements remain consistent across generative models, with only minor variations in recall and response length.
This demonstrates that the effectiveness of our approach stems from the amplification mechanism itself rather than any particular generative architecture, highlighting its robustness and generalizability.
\begin{figure}
    \centering
    \begin{minipage}{0.55\textwidth}
        \centering
        \renewcommand{\arraystretch}{1.2}
        \captionof{table}{Ablation study of the decoder layers used for latent editing used in our method.}
        \label{tab:effect-layer}
        \resizebox{.96\textwidth}{!}{
        \begin{tabular}{c cc ccccc}
        \toprule
        $l$ & CHAIR$_S$ & CHAIR$_I$ & Average & Length & Recall & $HAR_{\mathtt{@} 1}$ $\uparrow$\\
        \midrule
        \_ & 53.0 & 14.0 & 33.50 & 92.20 & 81.00 & 0.7300 \\
        3 & 47.8 & 12.7 & 30.25 & 92.44 & 80.05 & 0.7456\\
        6 & 51.6 & 14.3 & 32.95 & 92.16 & 81.50 & 0.7358\\
        9 & 52.4 & 14.4 & 33.40 & 92.99 & 81.82 & 0.7342\\
        12 & 50.8 & 13.8 & 32.30 & 91.78 & 80.77 & 0.7365\\
        15 & 51.0 & 14.1 & 32.55 & 92.38 & 80.45 & 0.7338\\
        18 & 51.6 & 14.0 & 32.80 & 91.47 & 80.51 & 0.7326\\
        21 & 51.8 & 13.7 & 32.75 & 91.78 & 81.36 & 0.7364\\
        24 & 51.0 & 13.4 & 32.20 & 91.72 & 80.45 & 0.7360\\
        \bottomrule
        \end{tabular}}
    \end{minipage}
    \hspace{5pt}
    \begin{minipage}{0.41\textwidth}
            \centering
            \renewcommand{\arraystretch}{1.2}
            \captionof{table}{Effect of our proposed hallucination amplification mechanism. We generate the reconstructed image using GPT-image~\citep{gptimage}.}
            \label{tab:effect-generative-model}
            \resizebox{.99\textwidth}{!}{
            \begin{tabular}{lccc}
            \toprule
            Metric & Baseline & FLUX.1-dev & GPT-image \\
            \midrule
            CHAIR$_S$ & 53.0 & 47.8 & 48.5 \\
            CHAIR$_I$ & 14.0 & 12.7 & 13.3 \\
            Average   & 33.50 & 30.25 & 30.90 \\
            Length    & 92.20 & 92.44 & 94.01 \\
            Recall    & 81.00 & 80.05 & 78.70 \\
            $HAR_{\mathtt{@}1}$ $\uparrow$ & 0.7300 & 0.7456 & 0.7359 \\
            \bottomrule
            \end{tabular}}
    \end{minipage}
\end{figure}
\section{Conclusion}
This paper proposes a training-free and self-supervised hallucination mitigation method for multimodal large language models, which leverages dual visual anchors to edit the hidden state in an end-to-end manner.
Our method can reduce hallucination at the object, attribute, and relation levels without sacrificing informativeness, while showing strong cross-architecture generalization. More importantly, our method achieves superior robustness on hallucination-free data and can serve as a plug-and-play module. 
We hope that our work inspires further exploration of training-free visual grounding techniques and serves as a practical baseline for building more faithful and reliable MLLMs.

\section{Ethic and reproducibility statement}
We introduce a novel method to mitigate hallucination problems in MLLMs, improving their safety and reliability for the community. The datasets (benchmarks) used for the evaluation and comparison of our method and baselines are publicly accessible, ensuring the transparency and reproducibility of our work. We will release our work to the community as soon as it is accepted, ensuring that our work is reproduced and grounded for other researchers and practitioners.

\bibliography{iclr2026/iclr2026_conference}
\bibliographystyle{iclr2026_conference}

\clearpage

\appendix
\section*{Appendix}\label{appendix}
\section{Discussion and Future Work}
\label{Discussion}
We discuss some potential limitations and future work for our method.
1). Following prior works~\citep{jiang2025interpreting}, we conduct latent editing in shallow layers and consistently use the third layer for simplicity.
Future work may explore more adaptive strategies for layer selection, such as dynamically identifying the most effective intervention points based on input characteristics or model states.
2). Our method relies on T2I generation to amplify hallucinations. While effective, the performance may be influenced by the fidelity of the T2I model, especially for fine-grained attributes such as color. Future research could integrate stronger generative backbones or develop selective filtering strategies to reduce noise introduced by imperfect reconstructions.
3). Our primary goal focuses on hallucination mitigation without hurting informativeness, yet this inevitably involves a trade-off.
Future work could explore more principled ways to balance this trade-off, for instance, through adaptive weighting schemes or context-aware editing strategies that tailor the strength of correction to the severity of hallucinations.

\section{AI Assistant Usage Statement}
During the preparation of this paper, we made moderate use of Large Language Models (LLMs) for text polishing and for assistance in non-core coding tasks. However, no LLMs were used in developing the ideas and determining the structure and content of this paper.

\section{Detailed Experimental settings}
\subsection{POPE}
We use the official benchmark from \citep{li2023evaluatingobjecthallucinationlarge} work, in which each of the three settings, random, popular, and adversarial, contains 3000 Question-Answer pairs. The meanings of these three settings as follows: \textbf{random} randomly selects non-existent objects to ask about; \textbf{popular} selects objects from the top half most frequently appearing in the entire image dataset but not present in the current querying image; \textbf{adversarial} first ranks all objects according to their co-occurrence frequencies with the ground-truth objects, and then selects the top-k frequent ones that do not exit in the image. The query template in the benchmark is "Is there a/an [object] in the image?"
\subsection{CHAIR}
To ensure the comparability of results, we use the 500 images sampled from MSCOCO by work \citep{zou2025looktwiceanswermemoryspace}. CHAIR is proposed to evaluate object hallucination in image captioning tasks, which have two variants:  per-instance (\textnormal{CHAIR\textsubscript{I}}) and per-sentence (\textnormal{CHAIR\textsubscript{s}}). 
These are defined as:
\begin{equation}
\begin{aligned}
        \text{CHAIR}_I = \frac{|\{\text{hallucinated objects}\}|}{|\{\text{all objects mentioned}\}|}, \;
     \text{CHAIR}_S = \frac{|\{\text{sentences with hallucinated objects}\}|}{|\{\text{all sentences}\}|}
\end{aligned}.
\end{equation}
As mentioned above, $HAR_{\beta}$ derived from $F_{\beta}$ score, $F_{\beta}$ formulates as follows:
\[
F_{\beta} = (1 + \beta^2) \cdot \frac{\text{Precision} \cdot \text{Recall}}{(\beta^2 \cdot \text{Precision}) + \text{Recall}}
\]

\[
\text{Precision} = \frac{TP}{TP + FP}, \quad 
\text{Recall} = \frac{TP}{TP + FN}
\]

\subsection{MME}
The official benchmark from MME \citep{fu2024mmecomprehensiveevaluationbenchmark} assesses model performance across 14 diverse vision-language subtasks covering both perception and cognition. Here we select \textbf{Existence}, \textbf{Count}, \textbf{Position}, and \textbf{Color} subtasks, which are most associated with the object and attribute hallucinations.

\section{Detailed Experimental results} 
Here we demonstrate two examples of the performance of our method on LLaVA-v1.5-7B. Figure \ref{fig:case3_study} shows the model’s performance in the Yes-or-No question is related to its image captioning. We can also observe this relation in Figure \ref{fig:case4_study} that the reconstructed image also contains a potted tree which does not exist. 
\begin{figure}[]
    \centering
    \includegraphics[width=.8\textwidth]{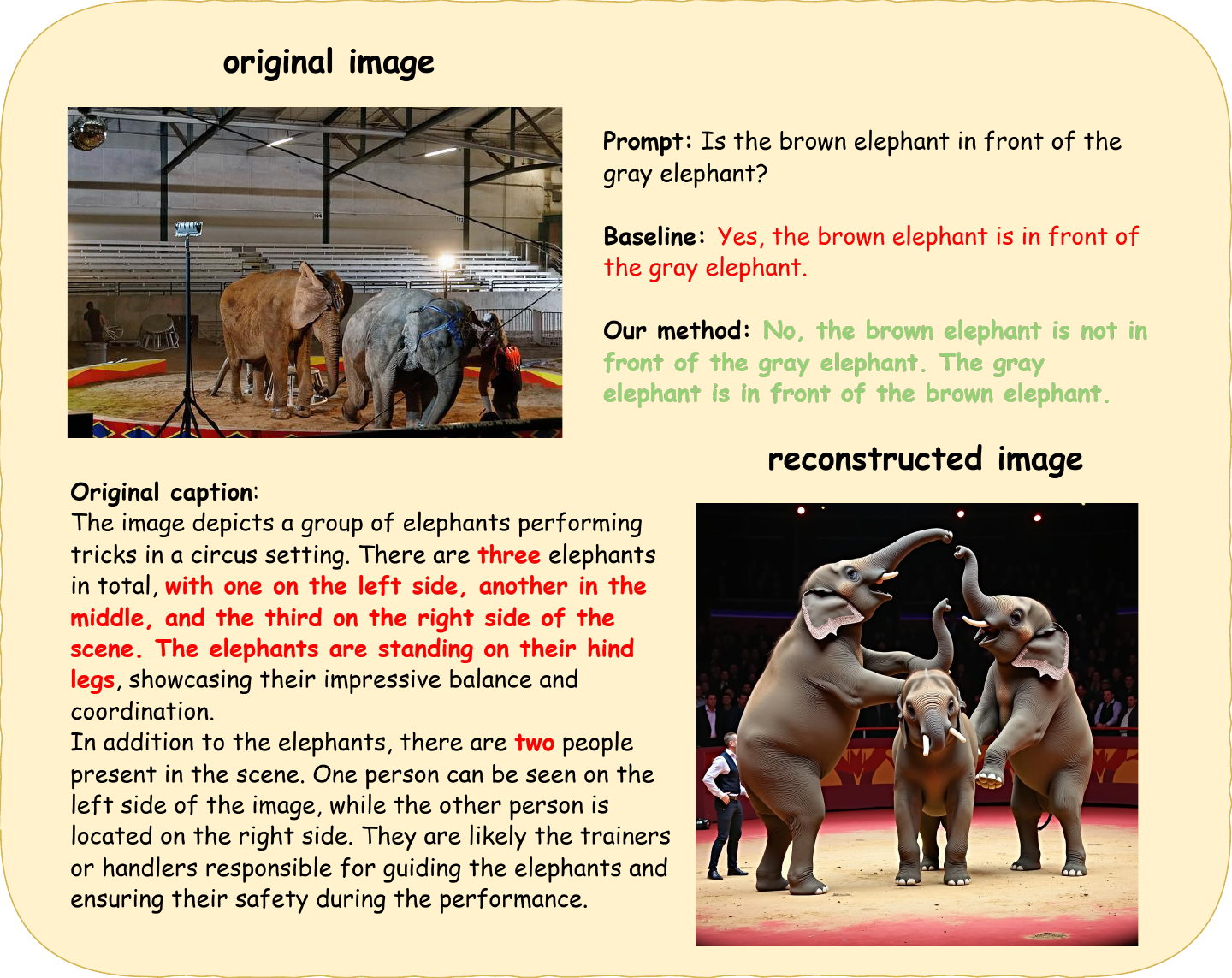}
    \caption{Illustration of comparison between baseline and our method generated by LLaVA-v1.5-7B on MME benchmark.}
    \label{fig:case3_study}
\end{figure}
\begin{figure}[]
    \centering
    \includegraphics[width=.8\textwidth]{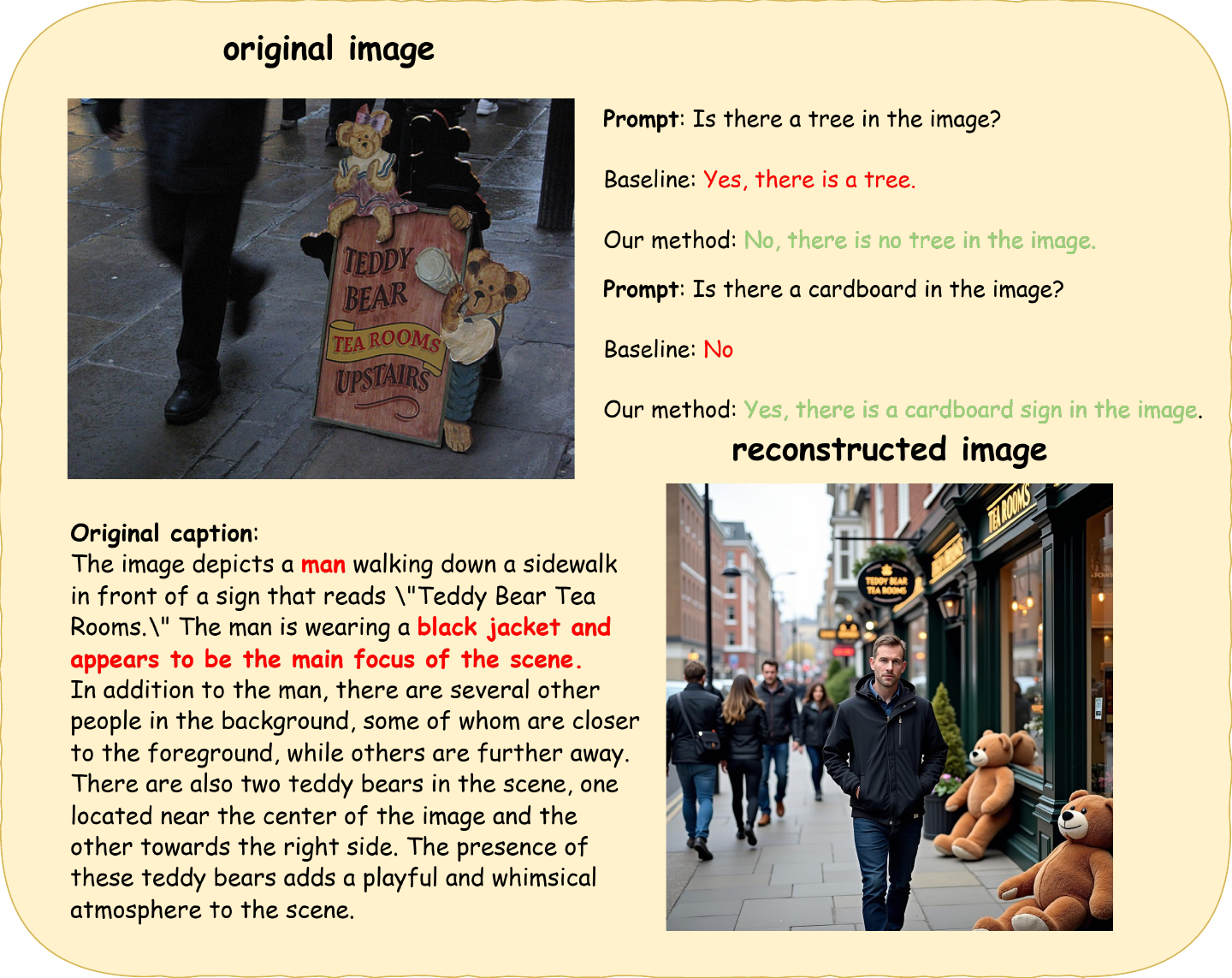}
    \caption{Illustration of comparison between baseline and our method generated by LLaVA-v1.5-7B on POPE benchmark. }
    \label{fig:case4_study}
\end{figure}

\begin{table}[t]
\centering
\renewcommand{\arraystretch}{1.2}
\caption{Effect of the choice of layer in latent editing across different architectures.}
\label{tab:layer-across-arch}
\resizebox{.9\textwidth}{!}{
\begin{tabular}{c| c cccccc}
\toprule
& $l$ &   CHAIR$_S$ & CHAIR$_I$ & Average & Length & Recall & $HAR_{\mathtt{@} 1}$ $\uparrow$\\
\midrule
\multirow{4}{*}{LLaVA-NEXT-7B}
& baseline  & 22.4 & 6.0 & 14.2 & 167.78 & 64.4 & 0.7358\\
& 3  & 6.6 & 2.9 & 4.75 & 119.51 & 60.2 & 0.7377\\
& 16 & 23.8 & 6.5 & 15.15 & 172.32 & 63.8 & 0.7283\\
& 24 & 23.2 & 6.2 & 14.7 & 174.81 & 64.4 & 0.7339\\
\midrule
\multirow{4}{*}{Cambrian-8B}
& baseline & 9.6 & 3.8 & 6.7 & 65.45 & 53.4 & 0.6792\\
& 3  & 8.0 & 2.9 & 5.45 & 67.43 & 52.2 & 0.6726\\
& 16 & 11.4 & 4.1 & 7.75 & 65.54 & 53.8 & 0.6796\\
& 26 & 11.4 & 4.5 & 7.95 & 66.36 & 52.4 & 0.6678\\
\midrule
\multirow{4}{*}{InstructBLIP-7B}
& baseline  & 57.4 & 15.5 & 36.45 & 98.08 & 74.7 & 0.6868\\
& 3  & 56.4 & 15.8 & 36.1 & 98.61 & 74.3 & 0.6871\\
& 16 & 54.2 & 15.4 & 34.8 & 97.93 & 75.1 & 0.6980\\
& 26 & 58.8 & 16.1 & 37.45 & 98.15 & 75.1 & 0.6825\\
\bottomrule
\end{tabular}}
\end{table}
\section{More Results on other MLLMs across more layers}
Table~\ref{tab:layer-across-arch} shows the performance of different models at various layers on CHAIR. We can observe that hallucinations are reduced in the shallow layers (3rd layer), while recall is well preserved. Table~\ref{tab:layer} demonstrates the CHAIR metric of our method applied to different layers of LLaVA-v1.5-7B. We can see that our method is effective when applied to most layers, and its best performance appears at the 3rd layer, possibly due to certain particularities, as other models also show strong hallucination mitigation at this layer.

\section{Robustness}
An intuitive idea is that the choice of $\alpha$ and $\beta$ should ideally depend on the input image and prompt, since the model exhibits different sensitivity to different inputs. At present, it is infeasible to determine the optimal parameter combination for a specific input. However, we can adopt parameter sampling strategies to explore better-performing values. Specifically, we sample from both uniform and Gaussian distributions(the average of $n$ samples), and evaluate the results under the conditions of $\alpha = \beta$ and $\alpha \neq \beta$, as shown in Table~\ref {tab:parameter_advice}. Note that a star indicates results obtained by repeating sampling five times and picking the best. Figure~\ref{fig:neq_uni} shows the distributions of $\alpha$ and $\beta$ corresponding to the starred results. We can observe that the parameter distributions are relatively uniform, which further demonstrates that our method does not require specific parameter designs. In fact, parameters randomly chosen within an appropriate range can effectively achieve hallucination mitigation.

\begin{figure}[t]
    \centering
    \includegraphics[width=1.0\textwidth]{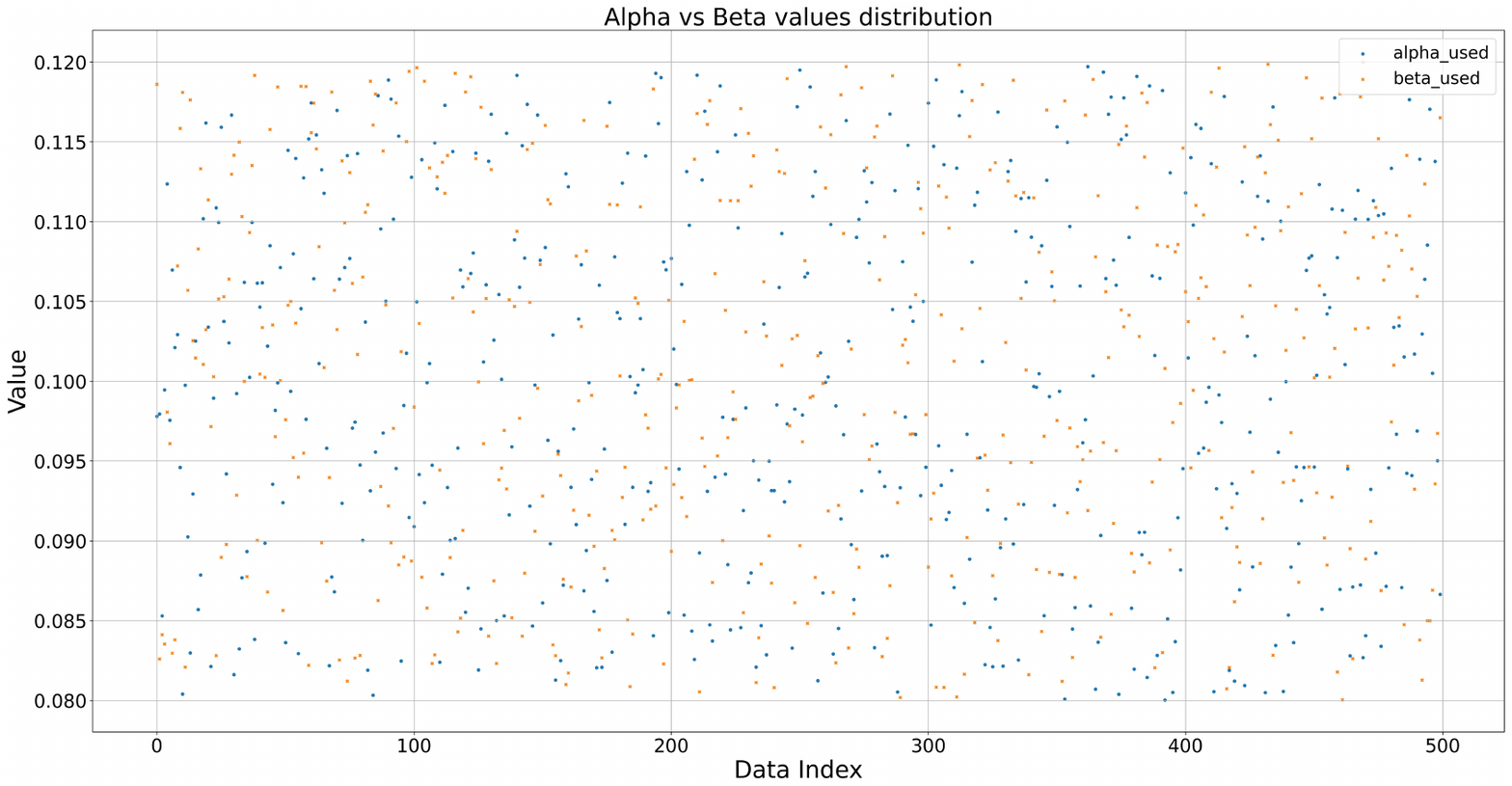}
    \caption{scatter plot of the $\alpha$ and $\beta$ obtained by randomly sampling five times from a uniform distribution and picking the best result, which shows that they are overall uniformly distributed within the range of 0.08 to 0.12. }
    \label{fig:neq_uni} 
\end{figure}

\begin{table}[t]
\centering
\caption{More parameter selection approaches on $\alpha$ and $\beta$. Here, $\mathcal{U}$ denotes sampling from a uniform distribution, $\mathcal{N}$ denotes sampling from a Gaussian distribution, and entries with a star ($^*$) indicate sampling five times and reporting the best result.}

\renewcommand{\arraystretch}{1.2}
\resizebox{.9\textwidth}{!}{
\begin{tabular}{c| cc cccccc}
\toprule
&\multirow{2}{*}{$\alpha$} &  \multirow{2}{*} {$\beta$}& \multicolumn{5}{c}{\textbf{CHAIR}} \\
\cmidrule(lr){4-9}
&  &  & CHAIR$_S$ & CHAIR$_I$ & Average & Length & Recall & $HAR_{\mathtt{@} 1}$ $\uparrow$\\
\midrule
\multirow{5}{*}{$\alpha = \beta$}
& 0 & 0 & 53.0 & 14.0 & 33.5 & 92.2 & 81.0 & 0.7304\\
& 0.1 & 0.1 & 47.8 & 12.65 & 30.23 & 92.4 & 80.05 & 0.7456\\
& $\mathcal{U}(0.08, 0.12)$ & $\mathcal{U}(0.08, 0.12)$ & 47.0 & 12.4 & 29.7 & 101.8 & 79.1 & 0.7444\\
& $\mathcal{N}(0.08, 0.12)$ & $\mathcal{N}(0.08, 0.12)$ & 47.4 & 12.2 & 29.8 & 100.9 & 79.9 & 0.7474\\
& $\mathcal{U}(0.08, 0.12)^{\star}$ & $\mathcal{U}(0.08, 0.12)^{\star}$ & 40.8 & 7.5 & 24.15 & 109.81 & 82.0 & 0.7881\\

\midrule
\multirow{3}{*}{$\alpha \neq \beta$}
& $\mathcal{U}(0.08, 0.12)$ & $\mathcal{U}(0.08, 0.12)$ & 48.6 & 14.4 & 31.5 & 90.7 & 78.2 & 0.7303\\
& $\mathcal{N}(0.08, 0.12)$ & $\mathcal{N}(0.08, 0.12)$ & 45.4 & 12.8 & 29.1 & 94.8 & 78.5 & 0.7451\\
& $\mathcal{N}(0.08, 0.12)^{\star}$ & $\mathcal{N}(0.08, 0.12)^{\star}$ & 40.6 & 7.1 & 23.85 & 114.0 & 80.5 & 0.7826\\
\bottomrule
\end{tabular}}
\label{tab:parameter_advice}
\end{table}

\begin{table}[]
\centering
\renewcommand{\arraystretch}{1.2}
\caption{Ablation results on the layers to edit}
\label{tab:layer}
\begin{tabular}{c cc ccccc}
\toprule
\multirow{2}{*}{$l$}& \multicolumn{5}{c}{\textbf{LLaVA-1.5}} \\
\cmidrule(lr){2-7}
 & CHAIR$_S$ & CHAIR$_I$ & Average & Length & Recall & $HAR_{\mathtt{@} 1}$ $\uparrow$\\
\midrule
1 & 53.0 & 14.81 & 33.91 & 91.97 & 79.46 & 0.7216\\
2 & 52.0 & 13.93 & 32.97 & 92.81 & 80.45 & 0.7313\\
3 & 47.8 & 12.65 & 30.23 & 92.44 & 80.05 & 0.7456\\
4 & 53.6 & 14.50 & 34.05 & 91.12 & 81.82 & 0.7303\\
5 & 52.0 & 14.29 & 33.15 & 92.54 & 81.89 & 0.7361\\
6 & 51.6 & 14.25 & 32.93 & 92.16 & 81.50 & 0.7358\\
7 & 50.8 & 14.24 & 32.52 & 90.85 & 80.77 & 0.7171\\
8 & 57.2 & 15.52 & 36.36 & 94.90 & 81.23 & 0.7137\\
9 & 52.4 & 14.43 & 33.42 & 92.99 & 81.82 & 0.7342\\
10 & 52.4 & 14.75 & 33.58 & 92.53 & 81.10 & 0.7303\\
11 & 51.2 & 14.27 & 32.74 & 91.77 & 81.17 & 0.7356\\
\midrule
12 & 50.8 & 13.84 & 32.32 & 91.78 & 80.77 & 0.7365\\
13 & 51.0 & 14.03 & 32.52 & 93.73 & 81.56 & 0.7385\\
14 & 53.0 & 14.30 & 33.65 & 93.33 & 80.38 & 0.7269\\
15 & 51.0 & 14.10 & 32.55 & 92.38 & 80.45 & 0.7338\\
16 & 50.4 & 13.84 & 32.12 & 91.61 & 81.10 & 0.7390\\
17 & 51.2 & 13.77 & 32.49 & 92.01 & 80.45 & 0.7341\\
18 & 51.6 & 13.99 & 32.80 & 91.47 & 80.51 & 0.7326\\
19 & 51.2 & 14.01 & 32.61 & 91.37 & 80.45 & 0.7334\\
20 & 50.4 & 13.92 & 32.16 & 91.62 & 80.97 & 0.7383\\
21 & 51.8 & 13.69 & 32.75 & 91.78 & 81.36 & 0.7364\\
\midrule
22 & 51.8 & 13.57 & 32.69 & 91.72 & 80.91 & 0.7349\\
23 & 51.4 & 13.77 & 32.59 & 92.05 & 80.64 & 0.7343\\
24 & 51.0 & 13.36 & 32.18 & 91.72 & 80.45 & 0.7360\\
25 & 52.2 & 14.11 & 33.16 & 91.49 & 81.43 & 0.7342\\
26 & 50.8 & 13.67 & 32.24 & 90.99 & 80.64 & 0.7364\\
27 & 53.6 & 14.15 & 33.88 & 92.13 & 81.23 & 0.7290\\
28 & 52.6 & 14.17 & 33.39 & 92.21 & 81.17 & 0.7317\\
29 & 52.8 & 14.14 & 33.47 & 92.22 & 81.30 & 0.7318\\
30 & 53.4 & 14.14 & 33.77 & 91.77 & 80.84 & 0.7281\\
31 & 53.2 & 13.85 & 33.53 & 91.84 & 81.30 & 0.7314\\
32 & 53.0 & 13.96 & 33.48 & 92.23 & 81.04 & 0.7307\\
\bottomrule
\end{tabular}

\end{table}

\end{document}